\documentclass{article}

\usepackage[utf8]{inputenc} 
\usepackage[T1]{fontenc}    
\usepackage{hyperref}       
\usepackage{url}            
\usepackage{booktabs}       
\usepackage{amsfonts}       
\usepackage{nicefrac}       
\usepackage{microtype}      
\usepackage{xcolor}         
\usepackage{wrapfig}

\usepackage{amsmath}
\usepackage{amssymb}
\usepackage{caption}
\usepackage{xcolor}
\usepackage[scale=2]{ccicons}
\usepackage{textpos}
\usepackage{url}
\usepackage{bbm}
\usepackage{balance}
\usepackage{graphicx}
\usepackage{todonotes}

\usepackage{natbib}
\setcitestyle{numbers,square}

\usepackage{subcaption}

\newcommand{\E}{\mathbb{E}}

\newcommand\indep{\protect\mathpalette{\protect\independenT}{\perp}}
\def\independenT#1#2{\mathrel{\rlap{$#1#2$}\mkern2mu{#1#2}}}

\newcommand{\CUd}{\textrm{Criteo Uplift} dataset }
\newcommand{\UMPlain}{Uplift Modeling }
\newcommand{\UM}{UM }
\newcommand{\ITEPlain}{Individual Treatment Effect }
\newcommand{\ITE}{ITE }

\title{A Large Scale Benchmark for Individual Treatment Effect Prediction and Uplift Modeling}

\author{
    Eustache Diemert$^{\clubsuit}$\thanks{contact author - e.diemert@criteo.com \  $\clubsuit$: Criteo AI Lab \  $\diamondsuit$: University Grenoble Alps \ $\spadesuit$: ENS Paris Saclay} \\
    Artem Betlei$^{\clubsuit,\diamondsuit}$ \\
    Christophe Renaudin$^{\clubsuit}$ \\ 
    Massih-Reza Amini$^{\diamondsuit}$ \\
    Théophane Gregoir$^{\clubsuit,\spadesuit}$ \\
    Thibaud Rahier$^{\clubsuit}$ \\
}

\begin{document}

\maketitle

\begin{abstract}
    Individual Treatment Effect (ITE) prediction is an important area of research in machine learning which aims at explaining and estimating the causal impact of an action at the granular level. It represents a problem of growing interest in multiple sectors of application such as healthcare, online advertising or socioeconomics. To foster research on this topic we release a publicly available collection of 13.9 million samples collected from several randomized control trials, scaling up previously available datasets by a healthy 210x factor. We provide details on the data collection and perform sanity checks to validate the use of this data for causal inference tasks. First, we formalize the task of uplift modeling (UM) that can be performed with this data, along with the relevant evaluation metrics. Then, we propose synthetic response surfaces and heterogeneous treatment assignment providing a general set-up for ITE prediction. Finally, we report experiments to validate key characteristics of the dataset leveraging its size to evaluate and compare $-$ with high statistical significance $-$ a selection of baseline UM and ITE prediction methods.
\end{abstract}

\section{Introduction}
The field of Causal Inference is at the heart of many scientific endeavors as researchers identify and quantify the causal effect of different factors in complex systems. More recently scientists studied the causal effect of treatments on individuals for applications such as personalized medicine, online advertising and targeted job training. Reproducible and more convincing research in this area came with the availability of benchmark datasets from real-world experiments, mainly using data collected in the 1980's for econometrics and medicine where sample sizes are relatively small. With the advent of the Internet cheap controlled experiments are now mainstream \cite{kohavi2009controlled} and make much larger scale studies possible (millions of samples being common). Unfortunately most of such datasets remain unavailable for open science as companies running these experiments keep them private. Such dataset size presents original challenges and is more representative of nowadays applications.

In this work we introduce a dataset of 14M instances containing data coming from online advertising company Criteo and having the following distinguishing characteristics: treatment imbalance, binary and continuous, anonymized features and low outcome rates. The dataset is available publicly on the \href{https://ailab.criteo.com/criteo-uplift-prediction-dataset/}{\color{blue}{Criteo AI Lab website}}.
With the release of this dataset we hope to bring modern scale, proper quality assurance and state-of-the-art documentation \cite{gebru2018datasheets} to the fields of Causal Inference and especially \ITEPlain{} (ITE) prediction and \UMPlain{} (UM).

\paragraph{Motivations for a new dataset}
Generally speaking the benefit of having more datasets at hand is that it enables to draw more robust conclusions when experimenting new methods as algorithms are run in a variety of settings with different characteristics. In particular when very few benchmarks are available in a very active field such as causal inference there is always a risk of "conceptual overfitting" as the research community tries to beat the state of the art on few datasets representing only a very narrow range of real applications. Moreover, the lack of public, realistic datasets sometimes prompts researchers to publish results on private data, thus making unreproducible claims.
Having large scale datasets is also a blessing for researchers in that they can run experiments on new methods with a greater chance to capture significant performance differences as the variance of metrics dwindles with size.

\paragraph{Related Works}
Pioneering work in \ITE prediction in the 2000's and 2010's such as \cite{hill2011bayesian} or \cite{shalit2017estimating} use two landmarks datasets: \textrm{IHDP}, a randomized infant development study (<1,000 samples) and \textrm{Jobs} \cite{lalonde1986evaluating} (approx. 3,000 samples), a randomized job training study, both collected in the 1980's. Very recent published work in \ITE prediction still use almost exclusively these datasets to day \cite{shalit2017estimating,louizos2017causal,yoon2018ganite}. Besides, in order to benchmark multiple methods, ACIC challenge \cite{Dorie2018} was held from 2016 to 2019 using IHDP dataset combined with diverse generation processes.
In the field of \UM a notable exception to private datasets is the Hillstrom study \cite{Hillstrom2008} (64,000 samples) collecting the sales results of an e-mail marketing campaign from the 2000's. Whilst a good improvement it is under documented and still lagging behind in terms of scale and covariates dimensionality (see Table \ref{table:data}) compared to modern datasets in other areas such as computer vision \cite{deng2009imagenet} or recommender systems \cite{bertin2011million} where rich features and the million samples landmark are common.

\begin{table}[!htb]
\centering
\caption{Summary of datasets characteristics.}
\vspace{2mm}
\resizebox{1.\textwidth}{!}{
\begin{tabular}{lccccc}
\toprule
{} Metric & \textrm{IHDP (Hill)} & \textrm{JOBS} & \textrm{IHDP (ACIC 2017)} & \textrm{HILLSTROM} & \textrm{CRITEO-UPLIFTv2} (ours) \\
\midrule
Size 		  & 747 & 3,212 & 4,302 & 42,693 & 13,979,592\\
Dimension 		  & 25 & 7 & 25 & 8 & 12\\
\hspace{20pt}  - Continuous & 6 & 3 & 6 & 2 & 4\\
\hspace{20pt}  - Binary & 19 & 4 & 19 & 3 & 0\\
\hspace{20pt}  - Multiple modalities & 0 & 0 & 0 & 3 & 8\\
Treatment Ratio & .19 & .09 & - & .50 & .85\\
\midrule
Avg. positive outcome (Visit / Conversion, \%) & - & 84.99 & - & 12.88 / 0.73 &  4.70 / 0.29\\
Relative Avg. Uplift (Visit / Conversion, \%) & - & -9.7 & - & 42.6 / 54.3 & 68.7 / 37.2\\
\bottomrule
\end{tabular}
}
\label{table:data}
\end{table}

\paragraph{Contribution 1: a large-scale, real-world dataset for \UM} this dataset brings a size increase of 2 orders of magnitude for \UM compared to established benchmarks.
In terms of covariate dimensionality it brings a much harder setting with some features having thousands of possible values, which is more representative of modern problems in Web applications. In the same line, the dataset is proposing challenges in both target and treatment imbalance with only a small portion of the users assigned to the control population and an overall low positive outcome rate.
The anonymization strategy employed on this dataset with features represented as hashed tokens is perhaps also a characteristic of future applications where privacy enhancing technologies are pervasive.
Finally, the dataset and its documentation are being maintained and improved over time as for instance a new version has been released one year after the initial one to correct a problem potentially impairing fair evaluation of models.

\paragraph{Contribution 2: a realistic benchmark for ITE prediction} For \ITE prediction specifically our dataset brings a size increase of 4 orders of magnitude compared to other benchmarks. Common usage is to design semi-synthetic experiments using real features and simulated outcomes defined by simple response surfaces (constant or exponential) \cite{hill2011bayesian}. We propose here additional, realistic surfaces that match observed patterns in real data and enrich the diversity of existing benchmarks.

\paragraph{Outline}
We discuss in Section \ref{sec:background} metrics and evaluation protocols for the \ITE prediction and \UM applications.
In the spirit of \cite{gebru2018datasheets} we detail the key elements from the datasheet of the dataset in Section \ref{sec:dataset}.
Then, in Section \ref{sec:semi-synth}, we introduce new synthetic response surfaces inspired by real observations that permit to use our large scale dataset for \ITE prediction task. Finally, we report experiments to validate key characteristics of the dataset in Section \ref{sec:experiments}.

\section{Background}
\label{sec:background}

Generally, both \ITE prediction and \UM have the common goal of determining how changing treatment affects outcome. However, a subtle difference between the two can be traced in the data generating process:
methods developed for \ITE prediction usually assume observational data and therefore additionally need to tackle the problem of 
selection bias. In \UM, on the other hand, experimental data 
from randomized control trial (RCT) is considered. 

\subsection{Notations/Framework}
To formalize the notions of \emph{Individual Treatment Effect} (ITE) and \emph{Uplift},
we introduce the following notations: we assume we have access to a dataset $\mathcal{D}=\{(x_i, t_i, y_i)\}_{1 \leq i \leq n}$ containing $n$ \textit{i.i.d.} realizations of random variables $X$ (covariates), $T$ (treatment) and $Y$ (outcome). More precisely, for a given individual $i$, $x_i$ belongs to some $d$-dimensional space $\mathcal{X}$ and contain the individual's features, $t_i \in \{0,1\}$ contains the individual's treatment status, \textit{i.e.} $t_i=1$ if $i$ received the treatment and $t_i=0$ otherwise, and $y_i \in \mathbb{R}$ is the value of the outcome corresponding to $i$.

Following the \emph{potential outcomes framework} \citep{rubin1974estimating}, each individual $i$ has two potential outcomes: $y_i(0)$ (if $i$ does not receive the treatment) and $y_i(1)$ (if $i$ receives the treatment) of which only $y_i = y_i(t_i)$ $-$ the \textbf{factual} outcome $-$ is observed. We denote $Y(0)$ and $Y(1)$ the underlying random variables.

For the interested reader, detailed reviews of the ITE prediction and UM fields are provided in \cite{zhang2020unified, Gutierrez2016CausalLiterature}. 

\subsection{\ITEPlain}
\paragraph{Definition}
The ITE of the individual $i$ is then given by the difference of its potential outcomes $y_i(1) - y_i(0)$, which are never both observed in the dataset $\mathcal{D}$. Since individuals are described by vectors of features $x$ we are rather interested by the \emph{conditional average treatment effect} (CATE) defined for any $x \in \mathcal{X}$ as:
\begin{equation}\label{eq:CATE-def}
    \tau(x) = \E[Y(1) - Y(0) | X = x].
\end{equation}
For any $x\in\mathcal{X}$, we define the \emph{response surfaces} $-$ or conditional potential outcomes $-$ as:
\begin{equation}\label{eq:defresponsesurface}
    \mu_0(x) = \E[Y(0)| X = x] \ \ \ \ \ \text{and}\ \ \ \ \ \mu_1(x) = \E[Y(1)| X = x],
\end{equation}
such that:
\begin{equation}\label{eq:CATE-def-respsurfaces}
    \tau(x) = \mu_1(x) - \mu_0(x).
\end{equation}

\paragraph{Metrics}
In order to evaluate ITE/CATE estimators, one may use an adapted version of the \emph{Mean-Squared Error} (MSE), namely the \emph{Precision in Estimation of Heterogeneous Effects} (PEHE) \citep{hill2011bayesian}, which is defined for a model $\hat{\tau}$ as:
\begin{equation}
    \epsilon_{PEHE}(\hat{\tau}) = \E\left[\Bigl(\tau(X) - \hat{\tau}(X)\Bigr)^2\right],
    \label{eq:PEHE}
\end{equation}
where the expectancy is taken with respect to the distributions of both $X$ and $\mathcal{D}$.

In a real-world dataset such as $\mathcal{D}$, one only observes the factual outcome $y_i(t_i)$ but not its \textbf{counterfactual} $y_i(1-t_i)$ $-$ this is referred to as the \emph{Fundamental Problem of Causal Inference} (FPCI) $-$ which prevents the access to the true value of the ITE and therefore makes the computation of the PEHE impossible. Most ITE works therefore experiment only on (semi-)synthetic data. 
    
A reasonable alternative to measure performance of ITE prediction without counterfactuals is to use the \emph{Policy Risk} metric \cite{shalit2017estimating}, which estimates the loss of targeting treatment according to the evaluated ITE model's predictions, at different population ratios.

\paragraph{Methods}
Standard approaches in ITE prediction include model-agnostic methods (meta-learners \citep{kunzel2019metalearners,nie2021quasi}, modified outcome methods \citep{athey2015machine} and their combinations \citep{kennedy2020optimal}) that implicitly predict the CATE, as well as tree-based approaches which are particularly suited for direct treatment effect estimation\citep{Athey2015,wager2018estimation}.
A prolific series of increasingly performing algorithm targeted towards ITE prediction 
have been proposed, using a variety of techniques to adjust for the covariate shift, such as generative adversarial nets \citep{yoon2018ganite}, auto-encoders \citep{louizos2017causal}, double machine learning \citep{chernozhukov2018double}, representation learning \citep{shalit2017estimating,zhang2020learning} or confounder balancing algorithms \cite{kuang2017estimating}.
Another recent trend is to study theoretical limits in ITE prediction and especially generalization bounds \citep{alaa2018limits}.

\paragraph{Evaluation protocol: from raw data to metric computation}
Individual Treatment Effect models are typically evaluated using semi-synthetic datasets $-$ which allow access the true ITE value and to the PEHE metric $-$ such as those proposed in \cite{hill2011bayesian, dorie2019automated}.
The two main design choices in such semi-synthetic datasets are the response surfaces for potential outcomes $\mu_0$ and $\mu_1$ defined in \eqref{eq:CATE-def-respsurfaces}, and the \emph{treatment assignment mechanism} or \emph{propensity score} \citep{curth2021doing} $p: x \mapsto \mathbb{P}(T=1|X=x)$. 
These choices have a direct impact on the difficulty of the corresponding ITE problem and on the type of model which will perform the best $-$ through aspects such as balance, overlap, alignment or treatment effect heterogeneity as described in \cite{dorie2019automated}.
For instance, in the popular IHDP semi-synthetic framework \citep{hill2011bayesian}, the chosen response surfaces are respectively exponential and linear for the control and treatment populations, and the treatment assignment is biased towards a selected proportion of the users. In the case of \cite{dorie2019automated}, several response surfaces with different levels of non-linearity, as well as several selection bias designs are tested. 

In less common scenarios when only real data are available and FPCI holds, one can evaluate and select ITE models either by using policy risk metric or by applying alternative model evaluation \citep{alaa2019validating} and model selection \citep{saito2020counterfactual} techniques. 

\subsection{\UMPlain}
\paragraph{Definition}
A practical oriented branch of the work on CATE estimation $-$ called uplift modeling \citep{Radcliffe2007} $-$ focuses on the 
RCT setting in which individuals are \textbf{randomly} split into a \emph{treatment} group and a \emph{control} group. This set-up circumvents the \emph{causal identification problem} (avoiding any bias in treatment assignment) and ensures the CATE (or uplift) is rightfully given by the difference between the following \textbf{conditional} expectations:
\begin{equation}
    u(x) = \E[Y|T=1, X=x] - \E[Y|T=0, X=x].
\end{equation}

\paragraph{Metrics}
When learning uplift models in a real-world setting, the PEHE cannot be computed as the ground-truth uplift is not observed (due to FPCI).
However, the RCT assumption allows the use of ranking-based metrics such as the Uplift Curve \cite{rzepakowski2010decision}, which considers a sorting of individuals according to the predicted uplift score in the descending order and measures the group uplift as a function of the percentage of top-ranked individuals. 
Intuitively, good uplift models will rank observations with larger ground-truth uplift higher and vice-versa (resembling the intuition of policy risk, where population ratio is also based on ranking of ITE scores and the users with higher predicted ITE induce smaller loss). 
There are several versions of uplift curves in the literature, differing in how the group uplift is calculated \cite{devriendt2020learning}. 
In that vein, the Area Under the Uplift Curve (AUUC) \cite{Jaskowski2012} represents the most popular metric in the community.

\paragraph{Methods}
UM methods are often overlapping with ITE prediction techniques or reinventing independently, as the former is subproblem of the latter. For instance, the two-model method is described both in ITE \citep{kunzel2019metalearners} and UM \citep{hansotia2002incremental} literature, the class variable transformation \citep{Jaskowski2012} is a particular case of \citep{athey2015machine}, and the tree-based method from the UM community \citep{rzepakowski2010decision, radcliffe2011real} only fractionally differ from ITE ones. However, there are methods specific to the UM
which either extend the two-model method by using certain representations \citep{betlei2018uplift} or aim at maximizing the AUUC directly \citep{Kuusisto2014,devriendt2020learning,betlei2021uplift}.

\paragraph{Evaluation Protocol}
Most of the time, different versions of AUUC are reported as performance measures in the UM literature. For both hyperparameter tuning and building of confidence intervals, popular supervised learning techniques are usually used, such as stratified cross-validation \citep{Jaskowski2012,devriendt2020learning} or random stratified sampling \citep{betlei2018uplift,devriendt2020learning}, also test set bound for AUUC \citep{betlei2021uplift} is applied, aiming to provide confidence intervals based on single test split.

\section{The CRITEO-UPLIFTv2 Dataset}
\label{sec:dataset}
Dataset is publicly available on \href{https://ailab.criteo.com/criteo-uplift-prediction-dataset/}{\color{blue}{Criteo's website}}; code used in the experiments and validation is available on \href{https://github.com/criteo-research/large-scale-ITE-UM-benchmark}{\color{blue}{Criteo Research Github}}.

\paragraph{Motivation and supporting organization} Criteo is a private company which has been committed to the advancement of reproducible Advertising Science for a long time with a track record of releasing 7 large scale datasets in the last 7 years, some of which became industry and academic standards. In general these datasets\footnote{see \url{https://ailab.criteo.com/ressources/} for the complete list of published datasets} are interesting in that they showcase problems at the frontier of current ML theory and practice, for instance high-dimensional, tera-scale prediction problems \cite{criteoterabyte}, counterfactual learning \cite{lefortier2016large} and credit assignment learning \cite{diemert2017attribution}.
In order to provide a realistic benchmark for \UM, the Criteo AI Lab built a dataset through a collection of online controlled experiments (A/B tests) on the Web in order to better study the individual effect of advertising on ad clicks and sales.
More precisely, the dataset is constructed by assembling data resulting from several \emph{incrementality tests}, a particular Randomized Control Trial procedure where a part of an online population is prevented from being targeted by advertising whilst the other part is subject to it. 

\begin{figure}
\centering
  \includegraphics[width=\linewidth]{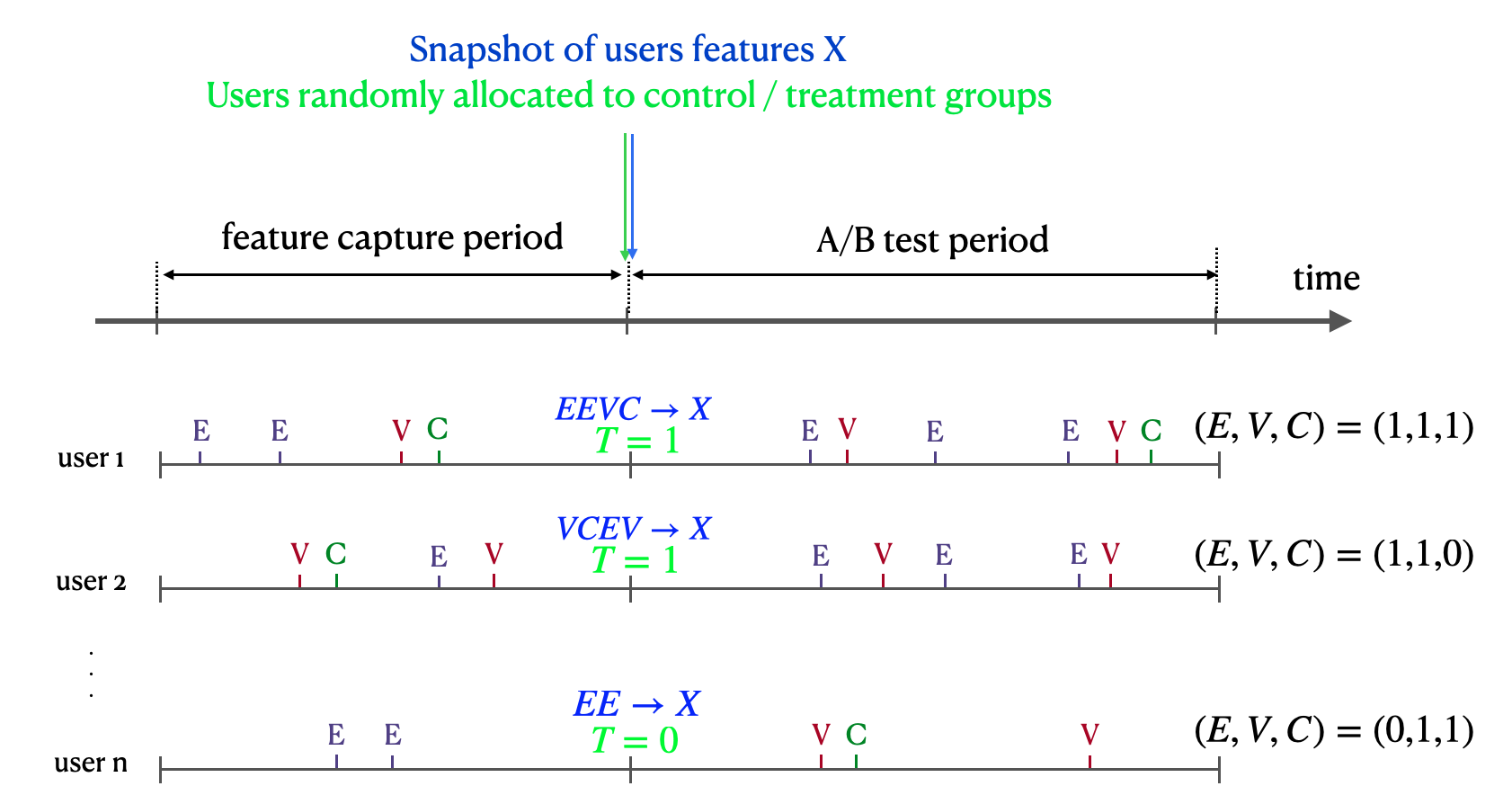}
 \caption{Data collection process illustration}
 \label{fig:datacolec}
\end{figure}

\paragraph{System description}
The system can be formally described by introducing the following variables: for a given user, $X$ contains their features, $T$ is the binary \emph{treatment} variable, such that $T=1$ for users in the treatment population and $T=0$ for user in the control population, and $E$, $V$ and $C$ are binary variables respectively indicating if the user has had at least one \emph{exposition} to advertisement, \emph{visit} on the website or \emph{conversion} during the A/B testing period (see Figure~\ref{fig:datacolec} for example timelines of such users).

\begin{figure}
    \centering
    \includegraphics[width=0.8\textwidth]{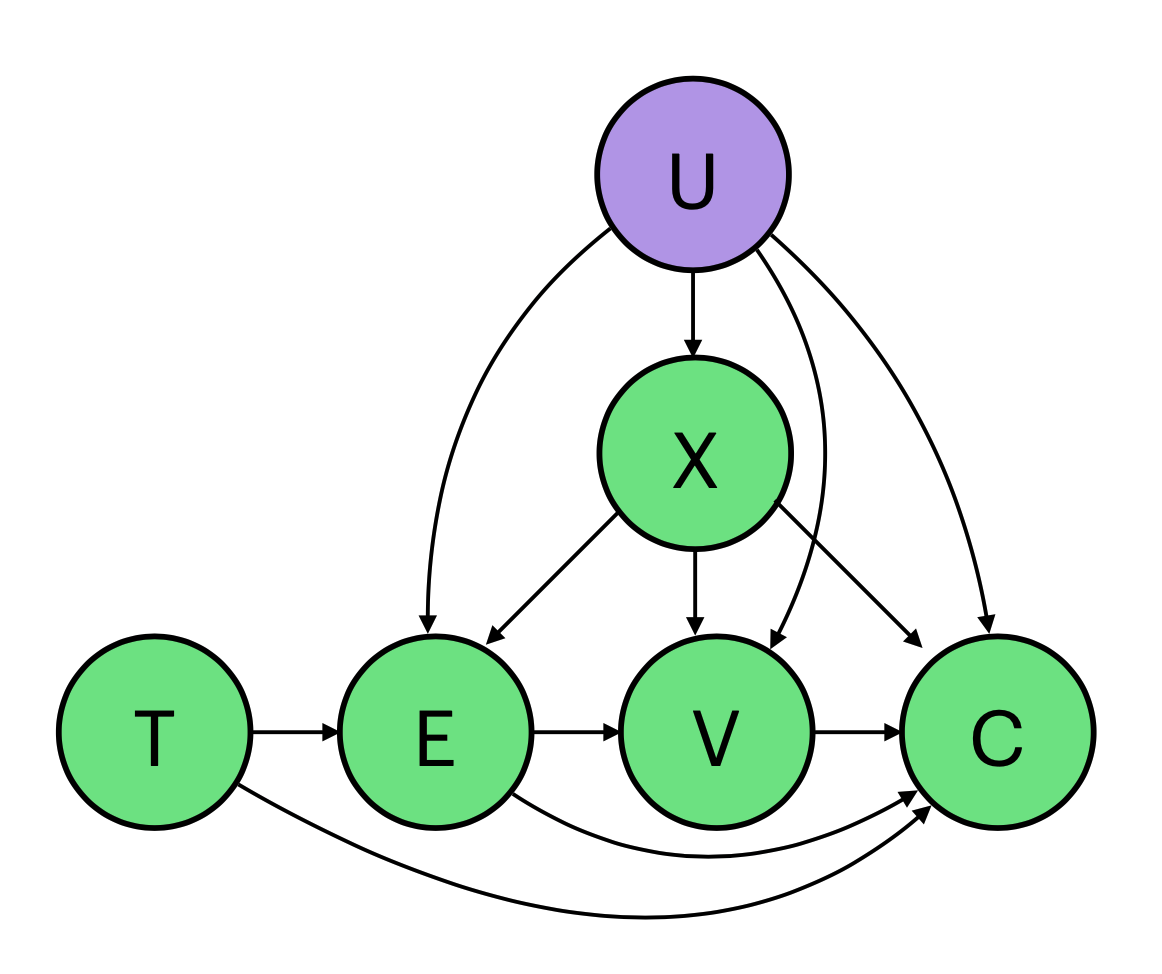}
    \caption{Causal graph of the online advertising system}
    \label{fig:scmv2}
\end{figure}

In Figure~\ref{fig:scmv2}, we present the underlying causal graph \citep{Pearl2000} associated to this system. It contains both conditional independence and causal relations. For example, we can deduce from the causal graph that the treatment ($T$) is independent on the user features ($X$), guaranteeing rightful causal effect identification, even though there exists unobserved variables ($U$) that might influence $E$, $V$ and $C$.

\paragraph{Structural aspects of the advertising application} There are a couple characteristics of the dataset that are structurally defined by the online advertising application. Firstly, the imbalance in treatment is explained by the reluctance to lose income when refraining to display ads in the control population. We believe that other applications may present the same challenge of not having sufficient experimentation budget to balance treatments in the whole population. Also, label imbalance is explained here by the relatively low conversion rate that is typical of online advertising applications. 

Finally, the variables respect the following constraints $-$ purely due to their definition in the online advertising context:
\begin{align*}
    T = 0 &\Rightarrow E = 0 &\text{ no exposition to ads in the control population}\\
    V = 0 &\Rightarrow C = 0 &\text{ no conversion can happen without a visit}
\end{align*}
The online advertising context suggests some additional assumptions that enable more efficient ITE prediction $-$ for example that the effect of $T$ on $C$ or $V$ is only impacted by $E$ \citep{rahier2021individual} $-$ which we will not detail further in this work.

\paragraph{Data collection}
As illustrated by Figure~\ref{fig:datacolec} users $-$ as identified by a browser cookie $-$ leave online traces through advertiser events such as website visits or product views \cite{kohavi2009controlled}.
For a given advertiser, at a pre-defined point in time a random draw assigns each user either to the treated or control population. The period before this assignment is used to capture user features (mostly related to prior user activity). The set of features was chosen so that it is predictive of subsequent user events and we can easily verify from a technical standpoint that they were all captured in the legit period. 
Once treatment is assigned users are then either subject to personalized advertising (if treated) or not (if in control) until the end of the data collection period. During the first 2 weeks after treatment assignment ad visits and online conversions on the advertiser website are logged. Then, features observed at treatment assignment times are joined with treatment prescription status, effective ad exposure and observed visits and conversion labels. Finally, the data for several such advertiser tests is merged to obtain the raw dataset.
Modern web platforms typically run numerous randomized experiments concurrently, yet users using these services are usually not fully aware of it. In our case we respected the \href{https://www.criteo.com/privacy/}{Criteo Privacy Policy} allowing users to opt out of the experiment at any point. The only drawback we can think of for users involved in the experiment was to avoid seeing ads, which is probably benign for most of us.

\paragraph{Anonymization}
To protect Criteo industrial assets and user privacy neither test advertiser provenance nor features names and meaning are disclosed. Moreover, feature values were hashed to a random vector space to make them practically impossible to recover while keeping their predictive power. Non-uniform negative sampling on labels has been performed so that the original incrementality level cannot be deduced while preserving a realistic, challenging benchmark.

\paragraph{Considerations to avoid temporal confounding} A particular characteristic of current advertising systems is that they target users dynamically based on observed interactions over time \cite{Berman2013}. This means that even in a randomized control trial (A/B test) interactions with the system influence subsequent ad exposure via adjustments of the bids based on user reactions. Notably, interactions after the first one are influenced both by the treatment and by previous interactions.
Possible solutions to avoid this temporal confounding include considering only the first interaction of a user during an A/B test or logging the user variables at the start of the test and observe the reward during the test. We have chosen the latter as it enforces logging of features at the same time for all users, minimizing chances to observe temporal shifts in their distribution for reasons like sales periods or production platform evolution.

\paragraph{Considerations to concatenate data from different tests}
The incrementality tests of different advertisers had different treatment ratios, meaning that the features as well as the uplift were correlated with the fact of being part of a given test. In other words the (unreleased) test id was a hidden confounder of the (features, labels) x treatment distribution. To allow for proper use for uplift and ITE prediction we needed that all instances in the final dataset were drawn i.i.d. from the same $P_{X,Y,T}$ distribution. If not, a prediction model could have had a good score by learning to predict which test an instance was coming from and exploiting the fact that some tests were heavily imbalanced in terms of treatment to guess if a treated or untreated positive was more likely. That would have defeated the purpose of the dataset to serve as a realistic benchmark for uplift or ITE modeling. To remedy that situation we sub-sampled all incrementality tests at the same, global treatment ratio. That way the dataset scale is preserved and the task is kept close to reality. This rebalancing is the key difference between the previous version (v1) and v2 of the dataset\footnote{v1 has been decommissioned and early users warned of that flaw} \cite{diemert2018large} and has been validated (see Section~\ref{sec:experiments_data_validation}).

\paragraph{Dataset description and analysis}
The final dataset (v2), henceforth referred to as CRITEO-UPLIFTv2, consists of 14M rows, each one representing a user with 12 features, a treatment indicator, an effective ad exposure indicator and 2 binary labels (visits and conversions).
The global treatment ratio is 85\%, meaning only a small portion of users where observed in the control population for which no advertising is performed by Criteo. It is typical that advertisers keep only a small control population as it costs them in potential revenue.
Positive labels mean the user visited/bought on the advertiser website during the test period (2 weeks).
Positive ad exposure means the user effectively saw an ad for the advertiser during the label collection period. 
Among the 12 variables, 4 are continuous and 8 are categorical with a large number of modalities. In order to evaluate the impact of each feature on the visit outcome $V$, a random forest model (formed by 100 estimators) is trained on each of the treatment and control population to predict $V$. Then, for each feature, Mean Decrease in Impurity (MDI) \cite{Breiman1984} is computed for both models and the corresponding average MDI is reported in Table \ref{table:features}.  According to this experiment, f0, f2, f8, f9 appear to drive $V$ significantly, while f1, f5, f11 have less influence.

\begin{table}[!htb]
\centering
\caption{Summary of CRITEO-UPLIFTv2 features: number of modalities (d) and importance (MDI)}
\label{table:features}
\vspace{2mm}
\resizebox{1.\textwidth}{!}{
\begin{tabular}{lrrrrrrrrrrrr}
\toprule
{}  & \textrm{f0} & \textrm{f1} & \textrm{f2} & \textrm{f3} & \textrm{f4} & \textrm{f5} & \textrm{f6} & \textrm{f7} & \textrm{f8} & \textrm{f9} & \textrm{f10} & \textrm{f11}\\
\midrule
\textbf{d} & \textrm{cont.} & 60 & \textrm{cont.} & 552 & 260 & 132 & 1,645 & \textrm{cont.} & 3,743 & 1,594 & \textrm{cont.} & 136\\
\textbf{MDI} ($10^{-3})$ & 110 & 3 & 297 & 36 & 20 & 7 & 64 & 18 & 218 & 170 & 46 & 10 \\
\bottomrule
\end{tabular}
}
\end{table}

\paragraph{Additional details and ethical considerations} More details are available in Sec. 2 of Supplementary; it includes the complete datasheet of the dataset as proposed by Gebru et al. \cite{gebru2018datasheets} which also covers detailed ethical and privacy questions. 

\section{CRITEO-ITE: generation of synthetic surfaces for ITE}
\label{sec:semi-synth}

In the spirit of \cite{hill2011bayesian}, we propose a class of synthetic response surfaces as well as method to design confounded treatment assignment, providing a semi-synthetic version of our dataset, named CRITEO-ITE, that can be used as a benchmark for ITE models evaluation.

\subsection{Response surfaces}
We add two classes of synthetic response surfaces for the CRITEO-ITE.

First, we reproduce the popular semi-synthetic setups from \cite{hill2011bayesian}. In the Case `A', constant treatment effect is generated by two linear response surfaces, namely $\mu_0(x) = x\beta$ and $\mu_1(x) = x\beta + 4$, where $\beta$ is a coefficient vector with each component sampled from the same multinomial distribution. Case `B' uses an exponential control response surface $\mu_0(x) = \exp((x + W)\beta)$ and a linear treatment response surface $\mu_1(x) = x\beta - \omega$, $W$ here is a fixed offset matrix and $\omega$ is a real number, adjusted so that the average treatment effect on the treated (ATT) is consistent with real-world measures. 

\begin{figure}[!htb]
  \centering
  \includegraphics[width=0.85\linewidth]{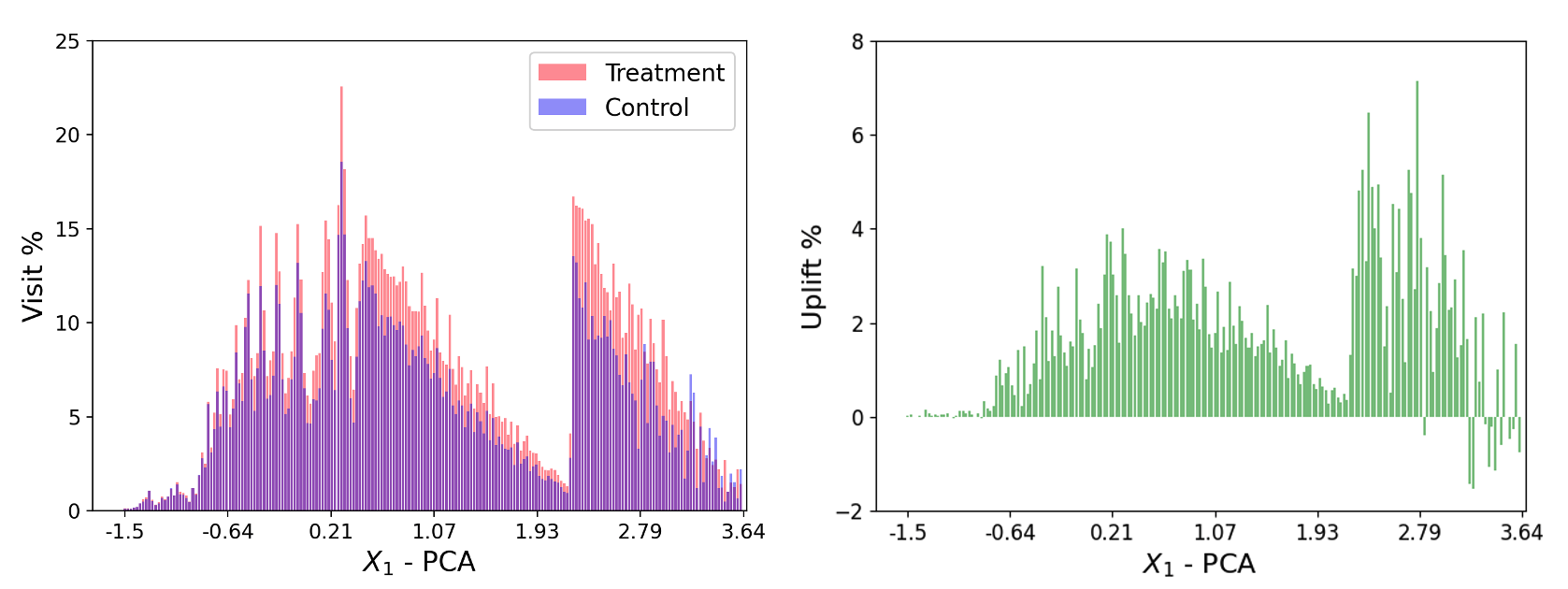}
  \caption{On the right, average uplift as a function of PCA first component computed on continuous features from CRITEO-UPLIFTv2 thanks to regular binning. On the left, average visit proportion computed with the same binning for control and treatment populations.}
  \label{fig:perf_visit}
\end{figure}

Second, we propose a novel \emph{multi-peaked} (non monotonous) class of response surfaces in the spirit of radial basis function interpolation $-$ inspired from observations made on projections of the real uplift surface (see Figure~\ref{fig:perf_visit}) $-$ which define both a novel and challenging ITE modeling problem.\\
Formally, we suppose that $\mathcal{X}$ is equipped with a norm $\vert\vert . \vert\vert$ and define, for $t\in\{0,1\}$ and $x\in\mathcal{X}$:
\begin{equation}\label{eq:mutlipeaked-def}
    \mu_t(x) = \sum_{c \in \mathcal{C}} w_{t,c}  \exp\left(- \frac{\vert\vert x - c \vert\vert^2}{2\sigma^2_c}\right),
\end{equation}
where $\mathcal{C}$ is a set of \emph{anchor points}, $\{w_{0,c}, w_{1,c}\}_{c \in \mathcal{C}}$ are the weights and $\{\sigma_{c}\}_{c \in \mathcal{C}}$ correspond to the width of influence associated to each of those points.

For any $x \in \mathcal{X}$, the associated CATE is therefore given by 
$$
\tau(x) = \sum_{c \in \mathcal{C}} \exp\left(- \frac{\vert\vert x - c \vert\vert^2}{2\sigma^2_c}\right) \left(w_{1,c} - w_{0,c}\right).
$$
If the distance between the different anchor points are large compared to the values of the $\sigma_c$s, the CATE of each $c \in \mathcal{C}$ is $\tau(c) \approx w_{1,c} - w_{0,c}$ and for any $x\in\mathcal{X}$, $\tau(x)$ is a weighted sum of the $\tau(c)$'s with weights exponentially decreasing with the ratios $\frac{\vert\vert x - c \vert\vert}{\sigma_c}$.
 
\subsection{Treatment assignment mechanism}
\label{sec:treatment_assignment}
To simulate an observational setting, we design an heterogeneous treatment assignment function $p: x \mapsto P(T=1|X=x)$ so that $T$ is confounded with the outcome $Y$ (note that the case where $p$ is constant corresponds to the RCT setting).

We propose a simple way to introduce treatment assignment bias by making $p$ depend on the component of $x$ which has the most predictive power with the outcome $Y$. Specifically, for a given small $\delta >0$ we define
\begin{equation}\label{eq:treatment-assignment}
    p(x) = (1 - 2\delta) \cdot \text{sigmoid}(\alpha^T x) + \delta,
\end{equation}
where $\alpha = (0, \dots, 0, 1, 0, \dots, 0)$ is a sparse $d-$dimensional vector for which the only nonzero component is the one which corresponds to the highest importance component of $x\in\mathcal{X}$ for the prediction of $\E[Y|X=x]$.

This choice of treatment assignment mechanism guarantees that the \emph{strong ignorability assumption} \citep{rosenbaum1983central} is met since $p(x) \in [\delta, 1-\delta]$ for all $x \in \mathcal{X}$, and that all confounders between $T$ and $Y$ are contained in $X$, ensuring that $(Y(1),Y(0)) \indep T \vert X$.


\section{Experiments}
\label{sec:experiments}
In this part, we will be presenting experiments conducted on CRITEO-UPLIFTv2 and CRITEO-ITE datasets. First, sanity checks are performed in order to validate a correct collection and creation of CRITEO-UPLIFTv2. Then, we provide experiments on CRITEO-UPLIFTv2 highlighting the impact of dataset size on separability for uplift modeling task. Finally, we provide a benchmark of ITE prediction methods on classical response surfaces and ours. 

\subsection{Dataset Validation}
\label{sec:experiments_data_validation}
We perform several sanity checks to verify properties of our dataset.

First, we ensure that the treatment is indeed independent of the features, defining the null hypothesis $H_0 \ : \ T \indep X$. A convenient way to verify this assumption is to perform a Classifier Two-Sample Test (C2ST) \citep{lopez2016revisiting}: under $H_0$, a classifier trained to predict $T$ from features $X$ should not perform better than a random classifier. Table \ref{tb:sanity_checks_indep} gives the result of the test, using log-loss to evaluate the classifiers performance. 
The empirical loss of the learned treatment classifier is not significantly different from the loss of a random classifier, which is reflected by a rather high p-value for the one-sided test: this does not allow to reject $H_0$ and validates that $T \indep X$.

Second, we make sure that logged features ($X$) are informative and relevant for predicting outcomes (visit $V$ and conversion $C$). This is not necessarily trivial as we sampled features that were technically easy to log and anonymized them. Table \ref{tb:sanity_checks_learn} presents the increase in performance of classifiers learned with visit and conversion as labels compared to baseline classifiers outputting a constant value (mean). We observe significative increase in performance indicating that features are indeed informative for the task.


\begin{table}[!htb]
\centering
    \centering
        \caption{Result of C2ST on treatment predictability (300 samples)}
        \vspace{5.5mm}
        \begin{tabular}{rrr}
        \toprule
         Median Random Loss &  Treatment Loss &  p-value \\
        \midrule
                    0.42269 &         0.42307 &  0.13667 \\
        \bottomrule
        \end{tabular}
    \label{tb:sanity_checks_indep}
\end{table}

\begin{table}[!htb]
   \centering
        \caption{Improvement over a dummy classifier for visit/conversion outcomes}
        \vspace{2mm}
        \begin{tabular}{lr}
        \toprule
        Outcome & Relative improvement (\%)\\
        \midrule
         visit & 34.74 \\
         conversion & 32.22 \\
        \bottomrule
        \end{tabular}
    \label{tb:sanity_checks_learn}
    \label{tb:sanity_checks}
\end{table}




\subsection{Uplift Modeling}
\label{sec:UM_task_exp}

\textbf{Features.} In order to reach a reasonable running time while conserving the great feature complexity of CRITEO-UPLIFTv2, the features used here are formed by the 4 initial continuous features and 100 projections on random vectors of the categorical features which are then one-hot encoded. 

\textbf{Target.} To train uplift models, both visits and conversions are available as the labels. However, as presented here, we suggest practitioners to model uplift primarily on visits in so far as conversion uplift signal appears to be too weak due to the high imbalance in the label.

\textbf{Metric.} We pick as a performance measure the "separate, relative" AUUC – evaluations of \citep{devriendt2020learning} concluded robustness of this version to treatment imbalance and its ability to capture the intended usage of uplift models to target future treatments. Confidence intervals are computed using AUUC test set bound \cite{betlei2020treatment}.

\textbf{Protocol.} The focus of this experiment is not on providing the best possible baseline but rather to highlight the fact that CRITEO-UPLIFTv2 is a meaningful alternative to existing UM datasets, scaling up in challenge while permitting to obtain statistical significance in the results. For this reason, we use 80\%/20\% train/test splits and AUUC performances were compared on test subsamples of proportional sizes to existing datasets, namely 1000 (IHDP), 5000 (Jobs), 50000 (Hillstrom), 1M and whole test data. Besides, the training set is used to tune the baseline models via grid search combined with stratified 5-fold cross-validation (to save both treatment and outcome imbalance).

\textbf{Models.} Four uplift models were used as a baselines: Two-Model (TM) \cite{hansotia2002incremental}, Class Variable Transformation (CVT) \cite{Jaskowski2012}, Modified Outcome Method (MOM) \cite{athey2015machine} and Shared Data Representation (SDR) \cite{betlei2018uplift}. For all models, regularization terms were tuned (see Supplementary).

\textbf{Results.} Figure \ref{fig:baselines_perf} represents results of the experiment. For the test sizes up to 1M, all presented methods are indistinguishable by their AUUC score as their confidence intervals overlap almost entirely. However, starting from 1M points onwards one can separate approaches and perform model selection. Hence it justifies the need for a large dataset for such a challenging task.

\begin{figure}[!htb]
\centering
\begin{subfigure}{.49\textwidth}
  \centering
  \includegraphics[width=\linewidth]{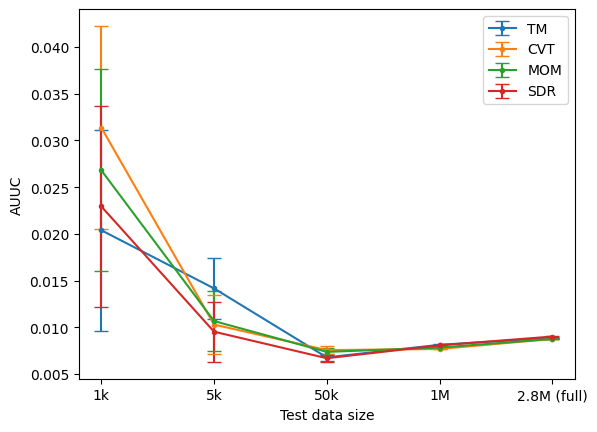}
  \label{fig:baselines_perf_all}
\end{subfigure}%
\begin{subfigure}{.49\textwidth}
  \centering
  \includegraphics[width=\linewidth]{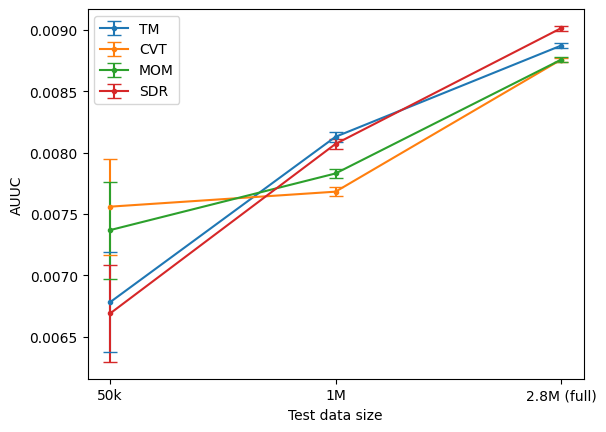}
  \label{fig:baselines_perf_2last}
\end{subfigure}%
\caption{Models separability on CRITEO-UPLIFTv2. }
\label{fig:baselines_perf}
\end{figure}


\subsection{ITE prediction}


Code for this benchmark experiment is available in the \href{https://github.com/criteo-research/large-scale-ITE-UM-benchmark}{\color{blue}{public repository}} and details including hyper-parameters are given in Supplementary.

\textbf{Features.} Features used for response surface generation and prediction are the 4 initial continuous and 5 random projections which are then one-hot encoded leading to a total dimensionality of 32.

\textbf{Target.} In order to test CRITEO-ITE, we performed a benchmark of ITE prediction baselines evaluated on 3 generation protocols presented in Section \ref{sec:semi-synth} : 
\begin{itemize}
    \setlength{\itemsep}{0pt}
    \setlength{\parskip}{0pt}
    \setlength{\topsep}{0pt}
    \item cases A and B from \cite{hill2011bayesian},
    \item our multi-peaked with 5 randomly selected anchor points, $\sigma_{c} = 1$ for all $c \in \mathcal{C}$ and $\{w_{0,c}, w_{1,c}\}_{c \in \mathcal{C}}$ drawn from  $\mathcal{U}(0, 1)$ then fitted to ensure $ATE \approx 4$ as for the other surfaces.
\end{itemize}
For the 3 types of surfaces, we defined the same treatment bias consisting in a sigmoid on the highest importance feature with $\delta = 0.01$ (see Section \ref{sec:treatment_assignment}). 

\textbf{Metric.} For each baseline, mean $\sqrt{\epsilon_{PEHE}}$ is reported with its standard deviation over 10 experiments.

\textbf{Protocol.} Following the concept of \cite{shalit2017estimating}, for each of the 3 generation protocols, 10 different realizations are generated. Then, for each realization, using a subsample of 100,000 points, baseline models are tuned thanks to a 5-fold cross-validation and then tested (with 50/50 train/test split).

\textbf{Models.} For this experiment, the baseline models include deep models (TARNet, CFRNet \cite{shalit2017estimating}) and meta-learners (T- and X-learner \cite{kunzel2019metalearners}, R-learner \cite{nie2021quasi}, and DR-learner \cite{kennedy2020optimal}) with Random Forest \cite{breiman2001random} as prediction models. 

\textbf{Results.} As illustrated by Table \ref{tab:ITE_exp}, performances differ from one type of surface to the other underlining the importance of developing a variety of responses in which our multi-peaked version can anchor. For example, although X-Learner outperforms R-Learner on surfaces from \cite{hill2011bayesian}, our multi-peaked generation highlights the contrary.






\begin{table}[!htb]
\centering
\caption{ITE prediction experiments on CRITEO-ITE. Mean $\sqrt{\epsilon_{PEHE}}$ performances are reported alongside their standard deviation. Best performance is in bold.}
\vspace{4mm}
\begin{tabular}{cccc}
\toprule
  ITE model           & Case A \cite{hill2011bayesian}        & Case B \cite{hill2011bayesian}    & Multi-peaked (ours) \\ 
\midrule
T-Learner  & $0.317 \pm 0.001$ & $1.890 \pm 0.001$  & $0.333 \pm 0.056$      \\
X-Learner  & $0.117 \pm 0.001$ & $1.883 \pm 0.002$  & $0.413 \pm 0.137$      \\
R-Learner  & $0.272 \pm 0.034$ & $11.668 \pm 2.897$ & $0.356 \pm 0.045$      \\
DR-Learner & \textbf{0.047 $\pm$ 0.005} & $2.510 \pm 0.342$  & $0.379 \pm 0.080$      \\
TARNet       & $0.104 \pm 0.001$ & $0.682 \pm 0.067$  & $0.195 \pm 0.044$      \\
CFRNet (MMD)       & $0.057 \pm 0.001$ & \textbf{0.239 $\pm$ 0.029}  & \textbf{0.152 $\pm$ 0.032}      \\ 
\bottomrule
\end{tabular}
\label{tab:ITE_exp}
\end{table}
\section{Conclusion}
We have highlighted the need for large scale benchmarks for causal inference tasks and released an open dataset, several orders of magnitude larger and more challenging than previously available. We have discussed the collection and sanity checks for its use in UM and ITE prediction. In particular we have shown that it enables research in uplift prediction with imbalanced treatment and response levels (e.g. visits) providing model separability due to its large size. We have also proposed semi-synthetic version of our dataset that can be used as a benchmark for ITE models evaluation. 

\newpage

\bibliographystyle{plain}
\bibliography{bibliography}

\newpage

\appendix

\section{Improvement on previous work}
\begin{figure}[h]
    \centering
    \includegraphics[scale=0.2]{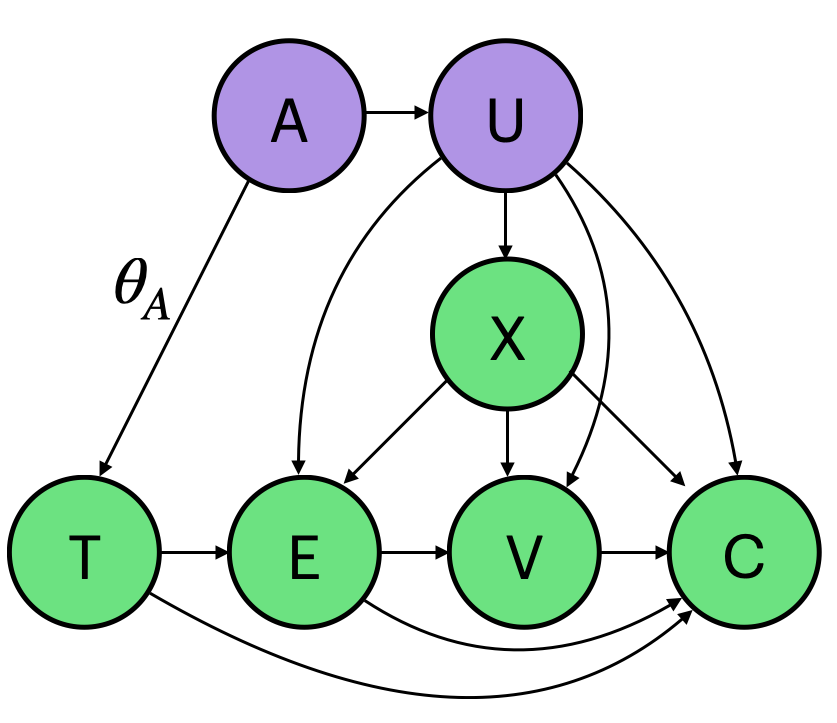}
    \caption{Causal graph displaying the relations between the different variables of interest in the 2018 version of the dataset \cite{diemert2018large}. Green and purple nodes respectively correspond to observed and unobserved variables. We see that because of the link between $A$ and $T$, there exists hidden confounding between $T$ and $C$.}
    \label{fig:causal_graph2}
\end{figure}

A first version of the CRITEO-UPLIFT dataset was released in 2018 \cite{diemert2018large}, in which the incrementality tests of different advertisers ($A$) had different treatment ratios $\theta_A$, implying that there existed a hidden confounder between $T$ and $C$, as shown in Figure~\ref{fig:causal_graph2}.

The new version of the dataset corrects this confounding problem when constructing the data, ensuring the treatment information ($T$) cannot be recovered from the features ($X$) $-$ and therefore that there are no hidden confounders between $T$ and other variables of interest. In addition, this version includes a new set of response surfaces (to be used in conjunction with the ITE dataset) that supplement previous ITE benchmarks.

\section{Experiment details}
 
For all experiments presented in Section 5, the code is available on the \href{https://github.com/criteo-research/large-scale-ITE-UM-benchmark}{\color{blue}{project repository}}. 

\subsection{Resources}
Experiments were conducted on Criteo internal cluster on instances with a RAM of 500Go and 46 CPUs available.

\subsection{Uplift Modeling}
The entire pipeline is described and available in the Jupyter Notebook "UM-Experiment.ipynb" for this experiment. 
For the model separability experiment (fig. 3), four uplift models were used as a baselines: Two-Model (TM) \cite{hansotia2002incremental}, Class Variable Transformation (CVT) \cite{Jaskowski2012}, Modified Outcome Method (MOM) \cite{athey2015machine} and Shared Data Representation (SDR) \cite{betlei2018uplift}. Particular prediction models (from scikit-learn \cite{scikit-learn}) and hyperparameter grids are the following:
\begin{itemize}
    \setlength{\itemsep}{0pt}
    \setlength{\parskip}{0pt}
    \setlength{\topsep}{0pt}
    \item TM, CVT – Logistic Regression, $l_2$ regularization term $C$ : $[1e^{0}, 1e^{2}, 1e^{4}, 1e^{6}, 1e^{8}]$
    \item MOM – Ridge, $l_2$ regularization term $\alpha$: $[1e^{-8}, 1e^{-6}, 1e^{-4}, 1e^{-2}, 1e^{0}]$
    \item SDR – Logistic Regression, $l_2$ regularization term $C$ : $[1, 10, 100, 1000]$, feature importance term $\lambda$: $[0.1, 1]$
\end{itemize}

\subsection{ITE prediction}
The entire pipeline is described and available in the Jupyter Notebook "ITE-Experiment.ipynb" for this experiment. 

TARNET and CFR were implemented thanks to TensorFlow \cite{tensorflow2015-whitepaper}. They were trained for 20 epochs during cross validation and finally for 100 epochs on the entire training set. Batch size was set to 128. 
For this two deep models, the following hyper parameters were tuned thanks to a randomized search :
\begin{itemize}
    \setlength{\itemsep}{0pt}
    \setlength{\parskip}{0pt}
    \setlength{\topsep}{0pt}
    \item number of layers : $[2, 3]$
    \item number of units per layer : $[32, 64]$
    \item regularization term : $[1e^{-4}, 1e^{-6}]$
    \item IPM regularization term (CFR only) : $[1e^{-2}, 1e^{-4}]$
\end{itemize}

Concerning meta-learners, models were partially implemented thanks to CausalML library \cite{chen2020causalml} (released with \href{https://github.com/uber/causalml/blob/df3830d798e3843364727ea8fdfa968261dc26e7/LICENSE}{\color{blue}{Apache License, Version 2.0}}). Meta-learners are using Random Forest Regressors from scikit-learn \cite{scikit-learn} which were tuned thanks to a randomized search with the following hyper parameters :
\begin{itemize}
    \setlength{\itemsep}{0pt}
    \setlength{\parskip}{0pt}
    \setlength{\topsep}{0pt}
    \item number of estimators : $[10, 20]$
    \item maximum depth in range : $[1,2,4,8]$
\end{itemize}

\section{Datasheet for the Criteo Uplift Dataset}

This section applies the methodology described in Gebru et al. 2018 \cite{gebru2018datasheets} to the presented dataset. It covers, among other details, the ethical and privacy aspects of the dataset.

\paragraph{Motivation}
\begin{itemize}
    \item \textit{For what purpose was the dataset created?} – Generally speaking the benefit of having more datasets at hand is that it enables to draw more robust conclusions when experimenting new methods as algorithms are run in a variety of settings with different characteristics. In particular when very few benchmarks are available in a very active field such as causal inference there is always a risk of "conceptual overfitting" as the research community tries to beat the state of the art on few datasets representing only a very narrow range of real applications. Moreover, the lack of public, realistic datasets prompts sometimes researchers to publish results on private data, thus making unreproducible claims.
Having large scale datasets is also a blessing for researchers in that they can run experiments on new methods with a greater chance to capture significant performance differences as the variance of metrics dwindles with size. Our dataset brings a size increase for \ITE of 4 orders of magnitude and 2 orders for \UM compared to established benchmarks.
In terms of covariate dimensionality it brings a much harder setting with some features having millions of possible values, which is more representative of modern problems in Web applications. In the same line the dataset is proposing challenges in both target and treatment imbalance with only a small portion of the users assigned to the control population and an overall low positive outcome rate.
The anonymization strategy employed on this dataset with features represented as hashed tokens is perhaps also a characteristic of future applications where privacy enhancing technologies are pervasive.
Finally, the dataset and its documentation is being maintained and improved over time as for instance a new version has been released one year after the initial one to correct a problem potentially impairing fair evaluation of models (see section 3 of paper for details).

For \ITE specifically, the common usage is to design semi-synthetic experiments using real features and simulated outcomes defined by simple response surfaces (constant or exponential) \cite{hill2011bayesian}; we additionally propose here more complex surfaces enriching the diversity of existing propositions.
    \item \textit{Who created the dataset (e.g., which team, research group) and on behalf of which entity (e.g., company, institution, organization)?} – Dataset was created by Causal Learning team of Criteo AI Lab.
    \item \textit{Who funded the creation of the dataset?} – Creation of the dataset was funded by Criteo.
\end{itemize}

\paragraph{Composition}
\begin{itemize}
    \item \textit{What do the instances that comprise the dataset represent (e.g., documents, photos, people, countries)?} – The instances that comprise the dataset represent people (internet users).
    \item \textit{How many instances are there in total (of each type, if appropriate)?} – There are 13979592 instances in total.
    \item \textit{Does the dataset contain all possible instances or is it a sample (not necessarily random) of instances from a larger set?} – Dataset contain all possible instances.
    \item \textit{What data does each instance consist of?} – Each instance consist of anonymised features related to prior user activity.
    \item \textit{Is there a label or target associated with each instance?} – There are two binary labels which are associated with each instance, namely facts of website visit and conversion.
    \item \textit{Is any information missing from individual instances?} – No.
    \item \textit{Are relationships between individual instances made explicit (e.g., users’ movie ratings, social network links)?} – No.
    \item \textit{Are there recommended data splits (e.g., training, development/validation, testing)?} – There are no specific data splits, however evaluation protocol is proposed, containing particular splitting approach.
    \item \textit{Are there any errors, sources of noise, or redundancies in the dataset?} – No error was spotted in the dataset for now. Errors could be coming from different sources: data collection error, user replication error, unintended visit due to a misclick.
    \item \textit{Is the dataset self-contained, or does it link to or otherwise rely on external resources (e.g., websites, tweets, other datasets)?} – Dataset is self-contained.
    \item \textit{Does the dataset contain data that might be considered confidential (e.g., data that is protected by legal privilege or by doctor- patient confidentiality, data that includes the content of individuals’ non-public communications)?} - Anonymisation and hashing were used to avoid this kind of issues. No confidential data is directly identifiable.
    \item \textit{Does the dataset contain data that, if viewed directly, might be offensive, insulting, threatening, or might otherwise cause anxiety?} – No.
    \item \textit{Does the dataset relate to people?} – Yes.
    \item \textit{Does the dataset identify any subpopulations (e.g., by age, gender)?} – No.
    \item \textit{Is it possible to identify individuals (i.e., one or more natural persons), either directly or indirectly (i.e., in combination with other data) from the dataset?} – No.
    \item \textit{Does the dataset contain data that might be considered sensitive in any way (e.g., data that reveals racial or ethnic origins, sexual orientations, religious beliefs, political opinions or union memberships, or locations; financial or health data; biometric or genetic data; forms of government identification, such as social security numbers; criminal history)?} – No.
\end{itemize}

\paragraph{Collection Process}
\begin{itemize}
    \item \textit{How was the data associated with each instance acquired?} – Users $-$ as identified by a browser cookie $-$ leave online traces through advertiser events such as website visits or product views \cite{kohavi2009controlled}.
For a given advertiser, at a pre-defined point in time a random draw assigns each user either to the treated or control population. The period before this assignment is used to capture user features (mostly related to prior user activity). The set of features was chosen so that it is predictive of subsequent user events and we can easily verify from a technical standpoint that they were all captured in the legit period. 
Once treatment is assigned users are then either subject to personalized advertising (if treated) or not (if in control) until the end of the data collection period. During the first 2 weeks after treatment assignment ad visits and online conversions on the advertiser website are logged. Then, features observed at treatment assignment times are joined with treatment prescription status, effective ad exposure and observed visits and conversion labels. Finally, the data for several such advertiser tests is merged to obtain the raw dataset.
    \item \textit{What mechanisms or procedures were used to collect the data (e.g., hardware apparatus or sensor, manual human curation, software program, software API)?} – Data is based on browser navigation (browsers send requests to Criteo datacenters).
    \item \textit{If the dataset is a sample from a larger set, what was the sampling strategy (e.g., deterministic, probabilistic with specific sampling probabilities)?} – N/A
    \item \textit{Who was involved in the data collection process (e.g., students, crowdworkers, contractors) and how were they compensated (e.g., how much were crowdworkers paid)?} – Only authors of the paper were involved in the data collection process.
    \item \textit{Over what timeframe was the data collected? Does this timeframe match the creation timeframe of the data associated with the instances (e.g., recent crawl of old news articles)?} – It took approximately 3 months to collect the dataset. 
    \item \textit{Were any ethical review processes conducted (e.g., by an institutional review board)?} – Yes. The collection and resulting dataset has been reviewed by an internal committee at Criteo focusing on legal, privacy and industrial property aspects.
    \item \textit{Does the dataset relate to people?} – Yes.
    \item \textit{Did you collect the data from the individuals in question directly, or obtain it via third parties or other sources (e.g., websites)?} – The data was obtained through browsers via cookies.
    \item \textit{Were the individuals in question notified about the data collection?} – Users were not directly notified, however they were asked whether to share their data via cookies or not. 
    \item \textit{Did the individuals in question consent to the collection and use of their data?} – Yes, according to \href{https://www.criteo.com/privacy/}{\color{blue}{Criteo Privacy Policy}}.
    \item \textit{If consent was obtained, were the consenting individuals provided with a mechanism to revoke their consent in the future or for certain uses?} – Yes, we respected the \href{https://www.criteo.com/privacy/}{\color{blue}{Criteo Privacy Policy}} allowing users to opt out of the experiment at any point.
    \item \textit{Has an analysis of the potential impact of the dataset and its use on data subjects (e.g., a data protection impact analysis) been conducted?} – Yes, neither test advertiser provenance nor features names and meaning are disclosed. Moreover, feature values were hashed to a random vector space to make them practically impossible to recover while keeping their predictive power. Non-uniform negative sampling on labels has been performed so that the original incrementality level cannot be deduced while preserving a realistic, challenging benchmark.
\end{itemize}

\paragraph{Preprocessing/cleaning/labeling}
\begin{itemize}
    \item \textit{Was any preprocessing/cleaning/labeling of the data done (e.g., discretization or bucketing, tokenization, part-of-speech tagging, SIFT feature extraction, removal of instances, processing of missing values)?} – Yes, feature values were hashed to a random vector space and non-uniform negative sampling on labels has been performed. 
    \item \textit{Was the “raw” data saved in addition to the preprocessed/cleaned/labeled data (e.g., to support unanticipated future uses)?} – No, to protect Criteo industrial assets and user privacy, “raw” data is not publicly available.
    \item \textit{Is the software used to preprocess/clean/label the instances available?} – No.
\end{itemize}

\paragraph{Uses}
\begin{itemize}
    \item \textit{Has the dataset been used for any tasks already?} – Dataset has been already used in several papers and preprints \citep{buzmakov2019comparison,betlei2021uplift,devriendt2020learning,gubela2019conversion,rahier2021individual,fernandez2021causal, gubelapreprint} and open source packages \citep{scikit_uplift,pyuplift} pertaining to UM or ITE prediction.
    \item \textit{Is there a repository that links to any or all papers or systems that use the dataset?} – No.
    \item \textit{What (other) tasks could the dataset be used for?} – Except the main tasks of UM and ITE prediction, dataset can be used for large-scale imbalanced binary/multi-label classification, multi-task or transfer learning.
    \item \textit{Is there anything about the composition of the dataset or the way it was collected and preprocessed/cleaned/labeled that might impact future uses?} – No. 
    \item \textit{Are there tasks for which the dataset should not be used?} – No.
\end{itemize}

\paragraph{Distribution}
\begin{itemize}
    \item \textit{Will the dataset be distributed to third parties outside of the entity (e.g., company, institution, organization) on behalf of which the dataset was created?} – Dataset is publicly available on \href{https://ailab.criteo.com/criteo-uplift-prediction-dataset/}{\color{blue}{Criteo Uplift Modeling Dataset}} page.
    \item \textit{How will the dataset will be distributed (e.g., tarball on website, API, GitHub)?} – Main dataset is distributed as a compressed (.gz) file that contains one comma-separated (.csv) file. Particular jupyter notebook (.ipynb) file is provided on our \href{https://github.com/criteo-research/large-scale-ITE-UM-benchmark}{\color{blue}{project repository}} to generate CRITEO-ITE data. 
    \item \textit{When will the dataset be distributed?} – Dataset is currently available.
    \item \textit{Will the dataset be distributed under a copyright or other intellectual property (IP) license, and/or under applicable terms of use (ToU)?} – Dataset is available under the \href{https://creativecommons.org/licenses/by-sa/4.0//}{\color{blue}{Creative Commons BY-SA 4.0}} which allows for copies and derivative work.
    \item \textit{Have any third parties imposed IP-based or other restrictions on the data associated with the instances?} – No.
    \item \textit{Do any export controls or other regulatory restrictions apply to the dataset or to individual instances?} – No.
\end{itemize}

\paragraph{Maintenance}
\begin{itemize}
    \item \textit{Who is supporting/hosting/maintaining the dataset?} – Criteo is supporting the dataset and is commited to make it available in the long run.
    \item \textit{How can the owner/curator/manager of the dataset be contacted
(e.g., email address)?} – by \href{e.diemert@criteo.com}{\color{blue}{email}}.
    \item \textit{Is there an erratum?} – Non uniformity of the incrementality level across advertisers caused the first version (v1) of the dataset to have a leak: uplift prediction could be artificially improved by differentiating advertisers using individual features (distribution of features being advertiser-dependent, see section 3 of the main paper). For this reason, an un-biased version (v2) of the dataset containing the same fields was released, which is available on \href{https://ailab.criteo.com/criteo-uplift-prediction-dataset/}{\color{blue}{Criteo Uplift Modeling Dataset}} page in "Erratum" section.
    \item \textit{Will the dataset be updated (e.g., to correct labeling errors, add
new instances, delete instances)?} – Dataset is already updated.
    \item \textit{If the dataset relates to people, are there applicable limits on the retention of the data associated with the instances (e.g., were individuals in question told that their data would be retained for a fixed period of time and then deleted)?} – No.
    \item \textit{Will older versions of the dataset continue to be supported/hosted/maintained?} – No, previous version has been decommissioned and early users warned of that flaw.
    \item \textit{If others want to extend/augment/build on/contribute to the dataset, is there a mechanism for them to do so?} – No.
\end{itemize}

\section{Author statement of responsibility}
Authors and Criteo bear all responsibility in case of violation of rights and confirmation of the data license.

\section{Research and Societal Impact} The \CUd will hopefully foster research in individual treatment effects modeling. We believe that methods developed in this field could have a large variation of impacts on society at large, depending on the way they are employed.
In the marketing domain personalized targeting of ads could be diversely experienced: from annoyance to serendipitous recommendations. In the health domain, targeting treatment to the individuals who would maximally benefit from it is the big promise of personalized medicine.
Finally we believe the nature of our contributions calls for a global positive impact. In general we hope it would help develop more reliable causal inference methods.

\end{document}